\documentclass{article}

     \usepackage[final]{ml4mols_2020}

\usepackage[utf8]{inputenc} 
\usepackage[T1]{fontenc}    
\usepackage{url}            
\usepackage{booktabs}       
\usepackage{amsfonts}       
\usepackage{nicefrac}       
\usepackage{microtype}      
\usepackage{graphicx} 
\usepackage{natbib}
\setcitestyle{square,numbers}
\usepackage{float}
\usepackage{chngcntr}
\usepackage{hyperref}       
\title{Attention-Based Learning on Molecular Ensembles}
\author{%
  Kangway V. Chuang\\
  Department of Pharmaceutical Chemistry\\
  Institute for Neurodegenerative Diseases\\
  University of California, San Francisco\\
  San Francisco, CA 94143\\
  \texttt{kangway.chuang@ucsf.edu} \\
  \And
  Michael J. Keiser \\
  Department of Pharmaceutical Chemistry\\
  Institute for Neurodegenerative Diseases\\
  University of California, San Francisco\\
  San Francisco, CA 94143\\
  \texttt{keiser@keiserlab.org} \\
}

\begin{document}

\maketitle

\begin{abstract}
The three-dimensional shape and conformation of small-molecule ligands are critical for biomolecular recognition, yet encoding 3D geometry has not improved ligand-based virtual screening approaches. We describe an end-to-end deep learning approach that operates directly on small-molecule conformational ensembles and identifies key conformational poses of small-molecules. Our networks leverage two levels of representation learning: 1) individual conformers are first encoded as spatial graphs using a graph neural network, and 2) sampled conformational ensembles are represented as sets using an attention mechanism to aggregate over individual instances. We demonstrate the feasibility of this approach on a simple task based on bidentate coordination of biaryl ligands, and show how attention-based pooling can elucidate key conformational poses in tasks based on molecular geometry. This work illustrates how set-based learning approaches may be further developed for small molecule-based virtual screening.

\end{abstract}

\section{Introduction}

Molecular shape and geometry are key for highly-specific biophysical recognition. Despite the critical importance of molecular conformation, ligand-based methods that incorporate three-dimensional features have had limited impact on virtual screening and drug discovery \citep{Maggiora2013-mm}. In contrast to structure-based methods, where multiple conformational poses of a molecule can be individually docked and scored against a protein pocket, methods based on ligand similarity present an inherent challenge. Drug-like molecules adopt diverse conformational shapes and orientations, yet for new systems, the relevant conformations are not known \textit{a priori}. Indeed, discovering key conformations and binding modes is often the goal of therapeutic discovery. Furthermore, this ambiguity in molecular representation introduces further challenges as most standard predictive models expect a single input.

\begin{figure}[t]
    \centering
    \includegraphics[width=13.80cm]{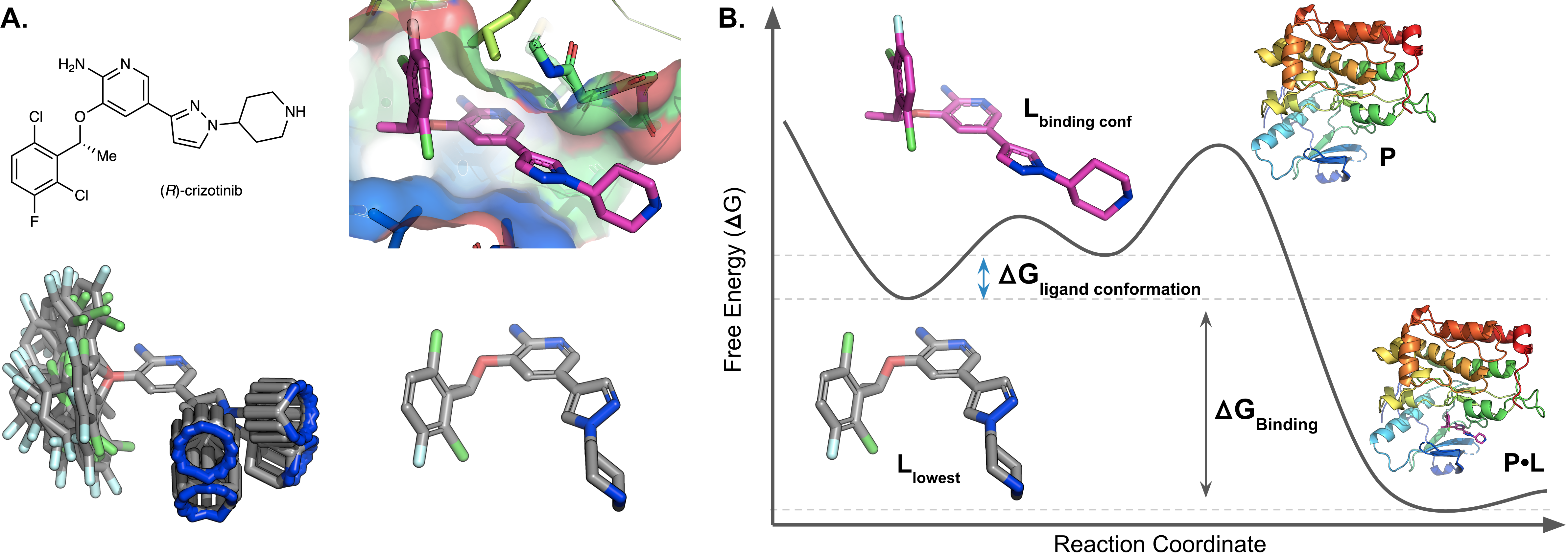}
    \caption{{\bf A.} Two-dimensional topological representations do not capture the inherent three-dimensional shape or dynamics of a molecule. Selecting a single low-energy conformer often does not reflect the salient geometric pose necessary for binding. {\bf B.} A simplified, hypothetical free energy landscape for protein-ligand binding (PDB 5AAB shown). Flexible molecules must traverse higher-energy conformations prior to ligation to the target protein.}
    \label{fig:figure_1}
\end{figure}

The challenges of input representation are  highlighted in the context of small-molecule ligand-protein binding (Figure \ref{fig:figure_1}). Whereas two-dimensional topological representations have been effective for similarity-based approaches, these fingerprints do not capture the spatial geometry of a small molecule, nor its diverse conformational ensemble (Figure \ref{fig:figure_1}A). Furthermore, although low-energy conformations can be generated readily, the lowest energy conformation of a drug often differs significantly from its bioactive pose (Figure \ref{fig:figure_1}B). \citet{Dietterich1997-lm} formalized this task as the multiple instance learning (MIL) problem, where a set of multiple input instances are mapped to a single output label.

Herein, we report an end-to-end deep multiple-instance learning approach that addresses the challenges of encoding molecular ensembles. We combine the expressive power of graph neural networks for molecular representation with an attention-based network for set aggregation to learn representations of conformational ensembles. We demonstrate on a new molecular dataset how this approach can be used to simultaneously provide good predictive performance while elucidating key conformational instances on tasks requiring molecular shape.

\section{Related Work}
\label{related_work}

\paragraph{Multiple-Instance Learning for Molecules} \citet{Dietterich1997-lm} first formalized the framework of multiple instance learning (MIL) motivated by small-molecule odor prediction. \citet{Fu2012-ey} and \citet{Zhao2013-xy} investigated support vector machine-based MIL  approaches for drug activity prediction by converting embedded sets into instances. Recently, \citet{Zankov2020-kr} reported a deep-MIL approach for drug-activity prediction using 3D-derived features and mean-pooling approach for set embedding. Our work differs from these seminal papers in using an end-to-end, differentiable graph neural network to simultaneously learn optimal feature extractors and a set aggregation function.

\paragraph{Attention-Based Pooling for Multiple-Instance Learning} 
\citet{Zaheer2017-hd} have characterized the requirements of permutation-invariant functions for set-based representations and illustrate their approach on set statistic estimation and anomaly detection. Our work builds off of \citet{Ilse2018-hz} who describe an attention-based, deep-multiple instance learning approach and demonstrate its application to diverse domains including histopathology.  \citet{Yan2018-ee} extended this method with a dynamic-routing based attention mechanism on similar tasks. Recently, \citet{Lee2019-qx} illustrate the strong performance of transformer-based networks for rich representation learning on sets.

\paragraph{Graph Neural Networks for Small Molecules} Our instance-level representations are motivated by recent work on graph neural networks for small molecules. Early work on differentiable molecular fingerprints by \citet{Duvenaud2015-au} and \citet{Kearnes2016-ab} illustrated performance gains vs standard 2D-topological graphs. The properties of individual conformers can be efficiently estimated by leveraging spatial features as illustrated by \citet{Gilmer2017-kv}, \citet{ Schutt2017-zf, Schutt2017-ir}, and \citet{Klicpera2020-qn}. In the context of small molecule bioactivity prediction,  PotentialNet by \citet{Feinberg2018-vs}   and ChemProp by \citet{Yang2019-wk} performed well across multiple benchmarks \citep{Wu2018-ai}. 

\section{Problem Formulation and Methods}

Whereas true biophysical systems are dynamic and dictated by complex enthalpic and entropic contributions, we greatly simplify protein-ligand binding problem to the static recognition of molecular geometry. In this framework, we invoke a variation of the molecular similarity principle \citep{Johnson1990-iu} and assume that two molecules are similar if they can adopt similar molecule geometries. We further assume that a molecule can be effectively represented by a set of discrete, sampled conformers. As a result, the similarity of two molecules can determined by their sets of of sampled conformers. In this domain, we aim to learn directly on molecular sets while identifying key conformational instances for similar molecules.

In standard ligand-based machine learning, each molecule serves as an input, $X$, with an accompanying label $y$. A multiple-instance learning approach \citep{Herrera2016-va} extends this supervised paradigm: each molecule is represented as a set of $K$ conformers, $X = \{\bf x_1, x_2, ... x_K\}$, where ${\bf x_K}$ denotes a single sampled conformer, with only a single set-level label $Y$. Although instances in the set are responsible for the overall set label, instance-level labels are not known \textit{a priori} (i.e. $y_k$ corresponding to $x_k$ are hidden or unknown). Importantly, molecules vary substantially in both size and conformational flexibility. Any representation of the set must therefore satisfy both permutation invariance (i.e. exchangeability of its elements) and handle variable-length set sizes. 

\begin{figure}[t]
    \centering
    \includegraphics[width=13.80cm]{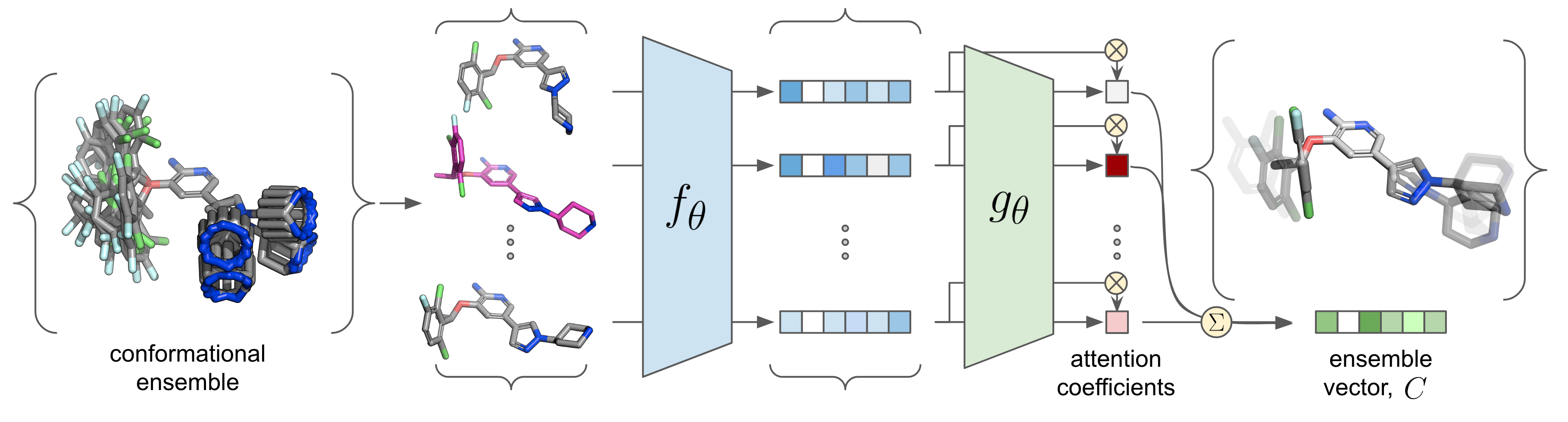}
    \caption{Our network encodes each conformer using a graph neural network, {$\bf f_\theta$}, as an instance featurizer, and uses an attention mechanism ${\bf g_\theta}$ to aggregate the embeddings from individual conformers. The resulting end-to-end differentiable network can be tied to both classification and regression tasks.}
    \label{fig:approach_overview}
\end{figure}

In this work, we describe an embedding-based approach with two levels of representation learning that consists of: 1) an instance featurizer, $f_{\theta}(x)$, that can expressively embed each conformer into a set of $K$ graph embeddings, $H = {{\bf \{h_1, h_2, ..., h_K}}\}$ and 2) a set-level aggregation function, $g_{\theta}(x)$, to pool instances into a fixed-length embedding (Figure \ref{fig:approach_overview}). Here, we represent all conformers as three-dimensional graphs, and our instance featurizer $f_{\theta}(x)$ is a graph neural network. We specifically use the edge-conditioned neural network as described by \citet{Gilmer2017-kv} and \citet{Simonovsky2017-tc}, with the corresponding message and update functions for each iteration, $t$:

\begin{equation}
    {\bf m_i}^{t+1} = \sum_{j \in N(i)}{\bf A(e_{ij}) \cdot h_j}^{t} \hspace{5pt} \textrm{with} \hspace{5pt} \
    {\bf h_i}^{t+1} = \textrm{GRU}({\bf m_i}^{t+1},{\bf h_i}^{t})
\end{equation}

The message from each graph neighbor ${\bf{m_i}}^{t+1}$ depends on the corresponding distance and edge-type $\bf{e_{ij}}$ through the learned weight matrix $\bf{A}$. Although any symmetric and differentiable function can serve as our set aggregator, $g_{\theta}(x)$, we opt to use an attention-based aggregation function \citep{Raffel2016-bc, Ilse2018-hz} for simplicity and interpretability. For a set of $K$ graph embeddings $H$:

\begin{equation}
     {\bf{c}} = \sum_{k=1}^K{\alpha_k {\bf{h_k}}}, \hspace{5pt} \textrm{with} \hspace{5pt}  {\alpha_k = \frac{e^{z_k}}{\sum_{j=1}^K{e^{z_j}}}}, \hspace{5pt}  \textrm{and} \hspace{5pt} {z_j = {\bf w^T} \tanh ({\bf{Vh_j^T})}}.
\end{equation}

The resulting context vector ${\bf c}$ is calculated as the expectation over each conformer instance, and represents the entire conformational ensemble. The parameters of the attention-layer, and the graph neural network are simultaneously optimized via backpropagation, and hence are trained specifically for the prediction task. From one perspective, the attention network learns a conformer-level energy function that is normalized across a softmax distribution, mimicking the classic Boltzmann distribution for molecular ensembles.

\section{Experiments and Results}
\label{Experiments and Results}

As a proof-of-concept model system, we constructed a small synthetic dataset of biaryl ligands to model protein-ligand binding (BIPY-MIL). We specifically analyze bidentate coordination modes, in which a \textit{s-cis} conformation of the 1,2-diamine is necessary for binding (see Appendix A). Each molecule in the dataset is considered positive if  \textit{at least one conformer} bears a 1,2-diamine substructure with a dihedral angle of 0 degrees. Note that the network is only given a set-level label. The key conformer instances are never revealed to the network and are withheld solely for evaluation.

We trained our models in a binary classification setting, and found comparable performance with random forest baselines known to work well in low data domains (100 and 500 training examples, Table \ref{results}). Here, the simple fingerprints derived from the molecular graph (ECFP4) alone can predict the set level; however, these 2D-baselines do not offer an interpretable method to identify key instances within each set.

\begin{figure}[b]
    \centering
    \includegraphics[width=13.8cm]{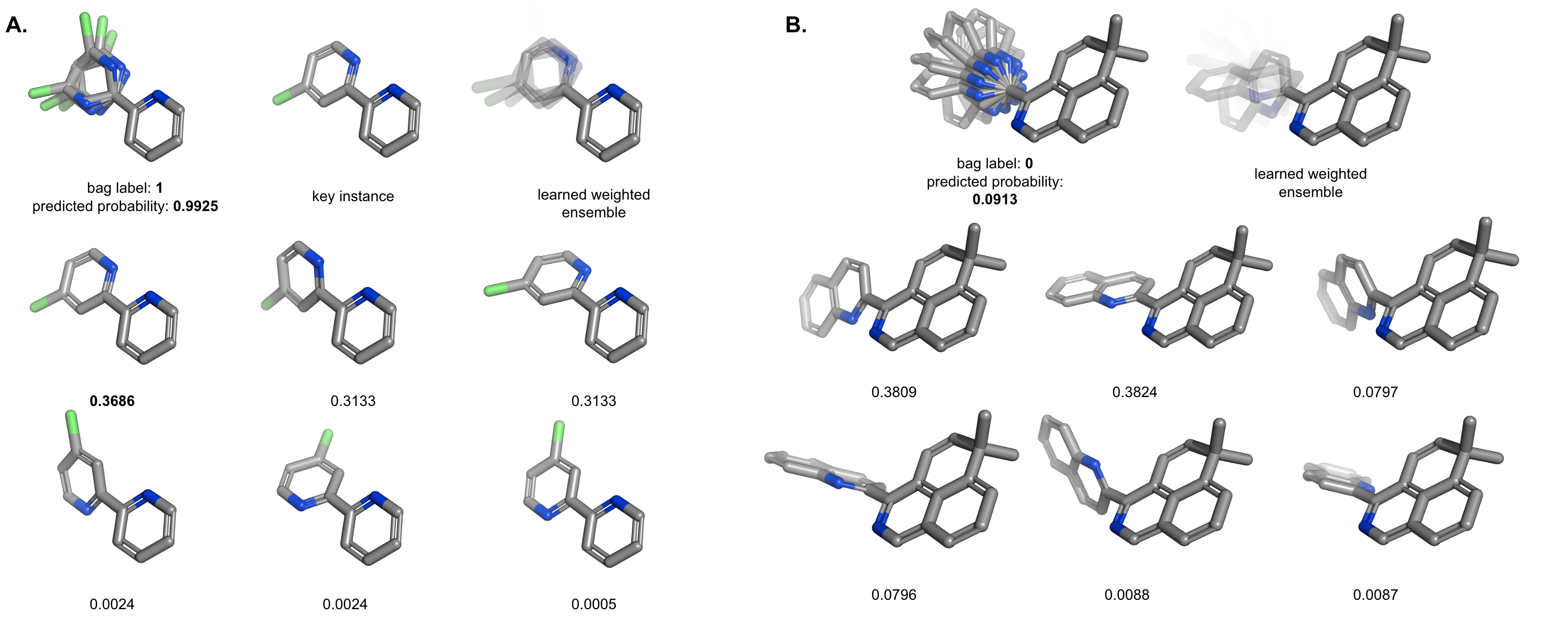}
    \caption{ Two examples from the BIPY-MIL dataset and their corresponding attention.  {\bf A.} A true positive example with high classification confidence contains a key instance where the N-C-C-N motif adopt a coplanar \textit{s-cis} conformation. {\bf B.} A true negative example that does not contain the coplanar pattern, but contains two close instances. In both cases, the attention coefficients are significant for conformations most similar to the key  instance (the instance that triggers the bag level).}
    \label{fig:attention}
\end{figure}

To understand the interpretability of this approach, we analyzed the attention coefficients for each conformer of each set, and used the rank order of the coefficients to compare against the hidden ground truth instance labels. Our model attributes its highest attention coefficient to a key instance in 80.7\% (Top-1) of all positive test cases. As a comparison, the lowest-energy pose only predicts the key instance in 3.8\% of cases, showing how encoding only a single lowest-energy conformer can create inject misleading bias. As depicted in Figure \ref{fig:attention}A, the attention coefficients aptly identify the key instance from a positive set ($\bf{1}$), with high coefficients for conformers that are similar, but slightly out of plane. Figure \ref{fig:attention}B shows a true negative example. The network simultaneously predicts a correct set label ($\bf{0}$), and although there is no key instance in the set, the attention coefficients remain highest for the two conformers with the smallest N-C-C-N dihedral angles. 

\begin{table}
  \caption{Classification results on the BIPY-MIL dataset.}
  \label{results}
  \centering
  \begin{tabular}{llllllll}
    \toprule
    \cmidrule(r){1-1}
    Model & $n$ &  Acc.  & AUROC & AUPRC & Top-1 & Top-5 & Top-10\\
    \midrule
    RF + ECFP4 & 100 & 0.901 & 0.962 & 0.917 & – & – & – \\
    RF + ECFP4 & 500 & {\bf0.960} & {\bf0.980} & {\bf0.967} & – & –  \\
    GNN + Attention & 100 & 0.886  & 0.956 & 0.918 & 0.628 & 0.840 & {\bf0.935} \\
    GNN + Attention & 500 & 0.928  & 0.976 & 0.957 & {\bf0.807} & {\bf0.885} & 0.923 \\
    Lowest Energy Pose & – & – & – & -  & 0.038  & 0.045 & 0.352 \\
    \bottomrule
  \end{tabular}
\end{table}

\section{Conclusions and Future Directions}
Our studies demonstrate the potential to learn on conformational ensembles for simple molecular systems. We have specifically illustrated how a simple attention mechanism can automatically retrieve key, driving instances from a large set of possible conformations using an embedding-based, multiple instance learning approach. These preliminary results lay a promising foundation toward learning more challenging molecular tasks, and we anticipate that additional adjustments in neural network architecture and attention mechanisms will enable this approach for complex tasks in small-molecule property and activity prediction. Our ongoing work is focused on further developing this approach and its applications toward real-world drug discovery.

\begin{ack}
We thank Laura Gunsalus for insightful discussion on graph neural networks and attention mechanisms. We also thank Will Connell, Dr. Jessica McKinley, and members of the Keiser laboratory for manuscript feedback and suggestions. OpenEye Scentific is gratefully acknowledged for an academic license to the OpenEye Toolkits, including OMEGA used for all conformer generation used in this paper. We are grateful to the Arnold and Mabel Beckman Foundation for generous support of this work through an Arnold O. Beckman Postdoctoral Fellowship in the Chemical Sciences to K. V. C., and the Chan Zuckerberg Initiative DAF, an advised fund of the Silicon Valley Community Foundation, for Grant 2018-191905 to M.J.K.
\end{ack}

\bibliography{bibliography}

\begin{thebibliography}{26}
\providecommand{\natexlab}[1]{#1}
\providecommand{\url}[1]{\texttt{#1}}
\expandafter\ifx\csname urlstyle\endcsname\relax
  \providecommand{\doi}[1]{doi: #1}\else
  \providecommand{\doi}{doi: \begingroup \urlstyle{rm}\Url}\fi

\bibitem[Maggiora et~al.(2013)Maggiora, Vogt, Stumpfe, and
  Bajorath]{Maggiora2013-mm}
Gerald Maggiora, Martin Vogt, Dagmar Stumpfe, and Jurgen Bajorath.
\newblock Molecular similarity in medicinal chemistry.
\newblock \emph{J. Med. Chem.}, 57\penalty0 (8):\penalty0 3186--3204, 2013.

\bibitem[Dietterich et~al.(1997)Dietterich, Lathrop, and
  Lozano-P{\'e}rez]{Dietterich1997-lm}
Thomas~G Dietterich, Richard~H Lathrop, and Tom{\'a}s Lozano-P{\'e}rez.
\newblock Solving the multiple instance problem with axis-parallel rectangles.
\newblock \emph{Artif. Intell.}, 89\penalty0 (1):\penalty0 31--71, January
  1997.

\bibitem[Fu et~al.(2012)Fu, Nan, Liu, Patel, Daga, Chen, Wilkins, and
  Doerksen]{Fu2012-ey}
Gang Fu, Xiaofei Nan, Haining Liu, Ronak~Y Patel, Pankaj~R Daga, Yixin Chen,
  Dawn~E Wilkins, and Robert~J Doerksen.
\newblock Implementation of multiple-instance learning in drug activity
  prediction.
\newblock \emph{BMC Bioinformatics}, 13 Suppl 15:\penalty0 S3, September 2012.

\bibitem[Zhao et~al.(2013)Zhao, Fu, Liu, Elokely, Doerksen, Chen, and
  Wilkins]{Zhao2013-xy}
Zhendong Zhao, Gang Fu, Sheng Liu, Khaled~M Elokely, Robert~J Doerksen, Yixin
  Chen, and Dawn~E Wilkins.
\newblock Drug activity prediction using multiple-instance learning via joint
  instance and feature selection.
\newblock \emph{BMC Bioinformatics}, 14 Suppl 14:\penalty0 S16, October 2013.

\bibitem[Zankov et~al.(2020)Zankov, Shevelev, Nikonenko, Polishchuk,
  Rakhimbekova, and Madzhidov]{Zankov2020-kr}
Dmitry~V Zankov, Maxim~D Shevelev, Alexandra~V Nikonenko, Pavel~G Polishchuk,
  Asima~I Rakhimbekova, and Timur~I Madzhidov.
\newblock Multi-instance learning for {Structure-Activity} modeling for
  molecular properties.
\newblock In \emph{Analysis of Images, Social Networks and Texts}, pages
  62--71. Springer International Publishing, 2020.

\bibitem[Zaheer et~al.(2017)Zaheer, Kottur, Ravanbakhsh, Poczos, Salakhutdinov,
  and Smola]{Zaheer2017-hd}
Manzil Zaheer, Satwik Kottur, Siamak Ravanbakhsh, Barnabas Poczos, Ruslan~R
  Salakhutdinov, and Alexander~J Smola.
\newblock Deep sets.
\newblock In I~Guyon, U~V Luxburg, S~Bengio, H~Wallach, R~Fergus,
  S~Vishwanathan, and R~Garnett, editors, \emph{Advances in Neural Information
  Processing Systems 30}, pages 3391--3401. Curran Associates, Inc., 2017.

\bibitem[Ilse et~al.(2018)Ilse, Tomczak, and Welling]{Ilse2018-hz}
Maximilian Ilse, Jakub~M Tomczak, and Max Welling.
\newblock Attention-based deep multiple instance learning.
\newblock In \emph{Proceedings of the International Conference on Machine
  Learning ({ICML)}, Stockholm, Sweden}, pages 10--15, 2018.

\bibitem[Yan et~al.(2018)Yan, Wang, Guo, Fang, Liu, and Huang]{Yan2018-ee}
Yongluan Yan, Xinggang Wang, Xiaojie Guo, Jiemin Fang, Wenyu Liu, and Junzhou
  Huang.
\newblock Deep multi-instance learning with dynamic pooling.
\newblock In Jun Zhu and Ichiro Takeuchi, editors, \emph{Proceedings of The
  10th Asian Conference on Machine Learning}, volume~95 of \emph{Proceedings of
  Machine Learning Research}, pages 662--677. PMLR, 2018.

\bibitem[Lee et~al.(2019)Lee, Lee, Kim, Kosiorek, Choi, and Teh]{Lee2019-qx}
Juho Lee, Yoonho Lee, Jungtaek Kim, Adam Kosiorek, Seungjin Choi, and Yee~Whye
  Teh.
\newblock Set transformer: A framework for attention-based
  {Permutation-Invariant} neural networks.
\newblock In Kamalika Chaudhuri and Ruslan Salakhutdinov, editors,
  \emph{Proceedings of the 36th International Conference on Machine Learning},
  volume~97 of \emph{Proceedings of Machine Learning Research}, pages
  3744--3753, Long Beach, California, USA, 2019. PMLR.

\bibitem[Duvenaud et~al.(2015)Duvenaud, Maclaurin, Iparraguirre, Bombarell,
  Hirzel, Aspuru-Guzik, and Adams]{Duvenaud2015-au}
David~K Duvenaud, Dougal Maclaurin, Jorge Iparraguirre, Rafael Bombarell,
  Timothy Hirzel, Alan Aspuru-Guzik, and Ryan~P Adams.
\newblock Convolutional networks on graphs for learning molecular fingerprints.
\newblock In C~Cortes, N~D Lawrence, D~D Lee, M~Sugiyama, and R~Garnett,
  editors, \emph{Advances in Neural Information Processing Systems 28}, pages
  2224--2232. Curran Associates, Inc., 2015.

\bibitem[Kearnes et~al.(2016)Kearnes, McCloskey, Berndl, Pande, and
  Riley]{Kearnes2016-ab}
Steven Kearnes, Kevin McCloskey, Marc Berndl, Vijay Pande, and Patrick Riley.
\newblock Molecular graph convolutions: moving beyond fingerprints.
\newblock \emph{J. Comput. Aided Mol. Des.}, 30\penalty0 (8):\penalty0
  595--608, August 2016.

\bibitem[Gilmer et~al.(2017)Gilmer, Schoenholz, Riley, Vinyals, and
  Dahl]{Gilmer2017-kv}
Justin Gilmer, Samuel~S Schoenholz, Patrick~F Riley, Oriol Vinyals, and
  George~E Dahl.
\newblock Neural message passing for quantum chemistry.
\newblock In \emph{Proceedings of the 34th International Conference on Machine
  Learning - Volume 70}, ICML'17, pages 1263--1272, Sydney, NSW, Australia,
  2017. JMLR.org.

\bibitem[Sch{\"u}tt et~al.(2017{\natexlab{a}})Sch{\"u}tt, Kindermans,
  Sauceda~Felix, Chmiela, Tkatchenko, and M{\"u}ller]{Schutt2017-zf}
Kristof Sch{\"u}tt, Pieter-Jan Kindermans, Huziel~Enoc Sauceda~Felix, Stefan
  Chmiela, Alexandre Tkatchenko, and Klaus-Robert M{\"u}ller.
\newblock {SchNet}: A continuous-filter convolutional neural network for
  modeling quantum interactions.
\newblock In I~Guyon, U~V Luxburg, S~Bengio, H~Wallach, R~Fergus,
  S~Vishwanathan, and R~Garnett, editors, \emph{Advances in Neural Information
  Processing Systems 30}, pages 991--1001. Curran Associates, Inc.,
  2017{\natexlab{a}}.

\bibitem[Sch{\"u}tt et~al.(2017{\natexlab{b}})Sch{\"u}tt, Arbabzadah, Chmiela,
  M{\"u}ller, and Tkatchenko]{Schutt2017-ir}
Kristof~T Sch{\"u}tt, Farhad Arbabzadah, Stefan Chmiela, Klaus~R M{\"u}ller,
  and Alexandre Tkatchenko.
\newblock Quantum-chemical insights from deep tensor neural networks.
\newblock \emph{Nat. Commun.}, 8:\penalty0 13890, January 2017{\natexlab{b}}.

\bibitem[Klicpera et~al.(2020)Klicpera, Gro{\ss}, and
  G{\"u}nnemann]{Klicpera2020-qn}
Johannes Klicpera, Janek Gro{\ss}, and Stephan G{\"u}nnemann.
\newblock Directional message passing for molecular graphs.
\newblock arXiv:2003.03123 [cs.LG], March 2020.

\bibitem[Feinberg et~al.(2018)Feinberg, Sur, Wu, Husic, Mai, Li, Sun, Yang,
  Ramsundar, and Pande]{Feinberg2018-vs}
Evan~N Feinberg, Debnil Sur, Zhenqin Wu, Brooke~E Husic, Huanghao Mai, Yang Li,
  Saisai Sun, Jianyi Yang, Bharath Ramsundar, and Vijay~S Pande.
\newblock {PotentialNet} for molecular property prediction.
\newblock \emph{ACS Cent. Sci.}, 4\penalty0 (11):\penalty0 1520--1530, November
  2018.

\bibitem[Yang et~al.(2019)Yang, Swanson, Jin, Coley, Eiden, Gao, Guzman-Perez,
  Hopper, Kelley, Mathea, Palmer, Settels, Jaakkola, Jensen, and
  Barzilay]{Yang2019-wk}
Kevin Yang, Kyle Swanson, Wengong Jin, Connor Coley, Philipp Eiden, Hua Gao,
  Angel Guzman-Perez, Timothy Hopper, Brian Kelley, Miriam Mathea, Andrew
  Palmer, Volker Settels, Tommi Jaakkola, Klavs Jensen, and Regina Barzilay.
\newblock Analyzing learned molecular representations for property prediction.
\newblock \emph{J. Chem. Inf. Model.}, 59\penalty0 (8):\penalty0 3370--3388,
  August 2019.

\bibitem[Wu et~al.(2018)Wu, Ramsundar, Feinberg, Gomes, Geniesse, Pappu,
  Leswing, and Pande]{Wu2018-ai}
Zhenqin Wu, Bharath Ramsundar, Evan~N Feinberg, Joseph Gomes, Caleb Geniesse,
  Aneesh~S Pappu, Karl Leswing, and Vijay Pande.
\newblock {MoleculeNet}: a benchmark for molecular machine learning.
\newblock \emph{Chem. Sci.}, 9\penalty0 (2):\penalty0 513--530, January 2018.

\bibitem[Johnson and Maggiora(1990)]{Johnson1990-iu}
M~A Johnson and G~M Maggiora.
\newblock \emph{Concepts and Applications of Molecular Similarity}.
\newblock John Wiley \& Sons, New York, 1990.

\bibitem[Herrera et~al.(2016)Herrera, Ventura, Bello, Cornelis, Zafra,
  S{\'a}nchez-Tarrag{\'o}, and Vluymans]{Herrera2016-va}
Francisco Herrera, Sebasti{\'a}n Ventura, Rafael Bello, Chris Cornelis, Amelia
  Zafra, D{\'a}nel S{\'a}nchez-Tarrag{\'o}, and Sarah Vluymans.
\newblock \emph{Multiple Instance Learning: Foundations and Algorithms}.
\newblock Springer, November 2016.

\bibitem[Simonovsky and Komodakis(2017)]{Simonovsky2017-tc}
Martin Simonovsky and Nikos Komodakis.
\newblock Dynamic {Edge-Conditioned} filters in convolutional neural networks
  on graphs.
\newblock \emph{2017 IEEE Conference on Computer Vision and Pattern Recognition
  (CVPR)}, 2017.

\bibitem[Raffel and Ellis(2016)]{Raffel2016-bc}
Colin Raffel and Daniel P~W Ellis.
\newblock {Feed-Forward} networks with attention can solve some {Long-Term}
  memory problems.
\newblock arXiv:1512.08756 [cs.LG], 2016.

\bibitem[Landrum(2006)]{Landrum2006-si}
Greg Landrum.
\newblock {RDKit}: Open-source cheminformatics.
\newblock \emph{https://rdkit.org}, 2006.

\bibitem[Hawkins and Nicholls(2012)]{Hawkins2012-al}
Paul C~D Hawkins and Anthony Nicholls.
\newblock Conformer generation with {OMEGA}: learning from the data set and the
  analysis of failures.
\newblock \emph{J. Chem. Inf. Model.}, 52\penalty0 (11):\penalty0 2919--2936,
  November 2012.

\bibitem[Paszke et~al.(2019)Paszke, Gross, Massa, Lerer, Bradbury, Chanan,
  Killeen, Lin, Gimelshein, Antiga, Desmaison, Kopf, Yang, DeVito, Raison,
  Tejani, Chilamkurthy, Steiner, Fang, Bai, and Chintala]{Paszke2019-fn}
Adam Paszke, Sam Gross, Francisco Massa, Adam Lerer, James Bradbury, Gregory
  Chanan, Trevor Killeen, Zeming Lin, Natalia Gimelshein, Luca Antiga, Alban
  Desmaison, Andreas Kopf, Edward Yang, Zachary DeVito, Martin Raison, Alykhan
  Tejani, Sasank Chilamkurthy, Benoit Steiner, Lu~Fang, Junjie Bai, and Soumith
  Chintala.
\newblock {PyTorch}: An imperative style, {High-Performance} deep learning
  library.
\newblock In H~Wallach, H~Larochelle, A~Beygelzimer, F~dAlch'e Buc, E~Fox, and
  R~Garnett, editors, \emph{Advances in Neural Information Processing Systems
  32}, pages 8026--8037. Curran Associates, Inc., 2019.

\bibitem[Fey and Lenssen(2019)]{Fey2019-ki}
Matthias Fey and Jan~Eric Lenssen.
\newblock Fast graph representation learning with {PyTorch} geometric.
\newblock arXiv:1903.02428 [cs.LG], March 2019.

\end{thebibliography}

\section{Appendix A: Dataset Creation and Details}
\begin{figure}[t]
    \centering
    \includegraphics[width=13.8cm]{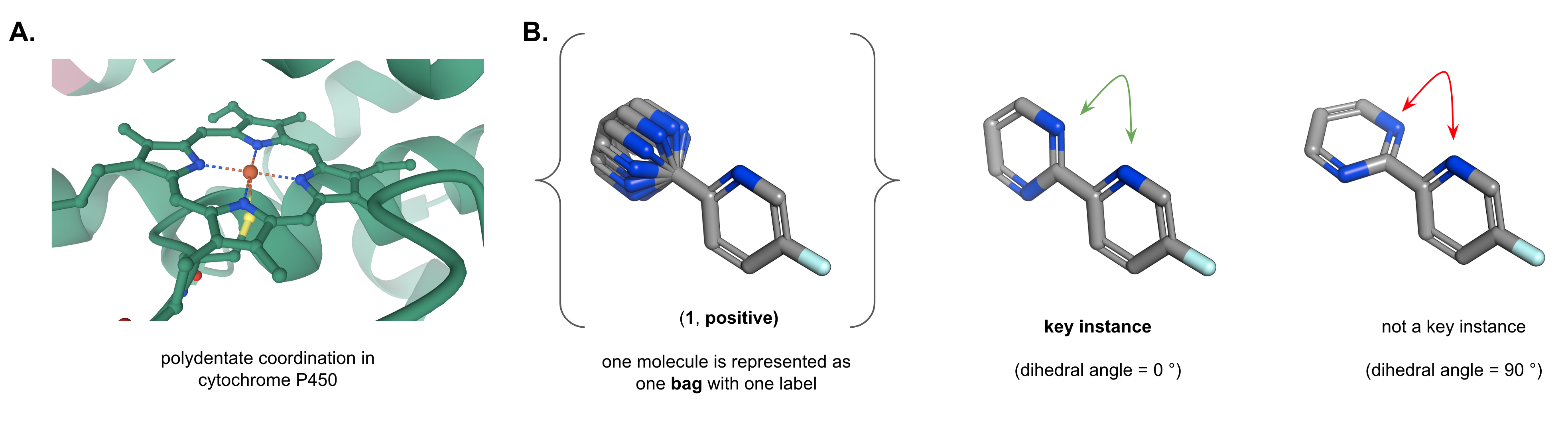}
    \caption{The BIPY MIL Task. {\bf A.} Metal coordination requires polydentate ligands. {\bf B.} We use a bipyridine scaffold to mimic cases where the coplanar nitrogens are able to coordinate a metal center, and use that as our key motif. }
    \label{fig:task}
\end{figure}

All molecules were analyzed and processed using the RDKit \citep{Landrum2006-si} and conformers generated using OpenEye OMEGA \citep{Hawkins2012-al}.

We constructed a toy dataset inspired by the bidentate coordination of substituted pyridines to a single attachment point (Figure \ref{fig:task}). In this task, the relevant binding mode requires the 1,2-diamine motif to adopt a dihedral angle of $0 ^{\circ}$. Despite the enthalpic benefit of bidentate bonding, the \textit{s-cis} conformation of bipyridine is typically not the lowest energy conformer. We enumerated a small synthetic library of biaryl ligands varying in substitution pattern and conformational rigidity. For each small-molecule, we used OpenEye OMEGA to generate up to 30 distinct conformations.

As labels, we analyzed each conformational ensemble for a \textit{s-cis} diamine motif, dictated by a dihedral angle of $< 1 ^\circ$. Each molecule (set of conformers) is assigned a a \textbf{0/1} binary label if at least one conformer in the set is able to adopt a coplanar bipyridine. The final dataset consists of 1,157 molecules, 15,959 conformers, with 398 positives and 759 negatives. A random sampling of representative examples are shown in Appendix Figure \ref{fig:bipys}, and additional summary statistics described in Appendix Table 2.

\begin{figure}[b]
    \centering
    \includegraphics[width=13.8cm]{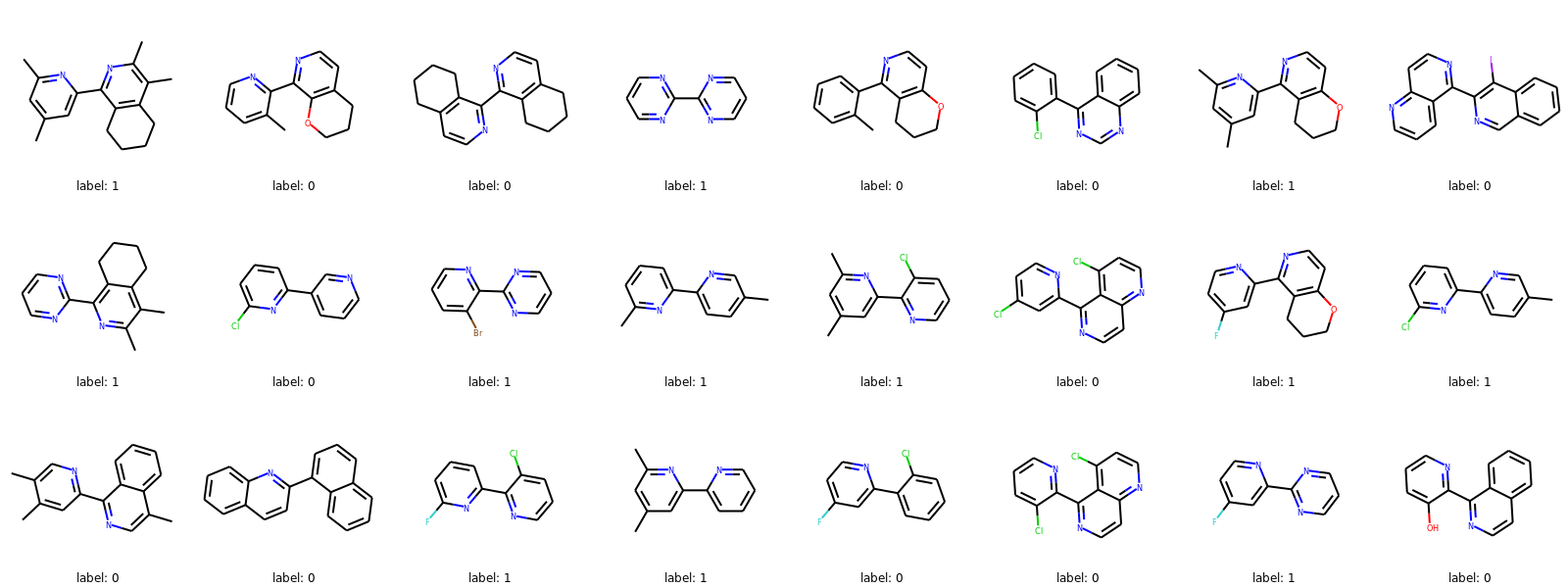}
    \caption{A random sample of molecules used in the BIPY dataset and their associated labels.}
    \label{fig:bipys}
\end{figure}

\begin{table}
  \caption{Detailed Statistics of the BIPY-MIL Dataset (n = 1157).}
  \label{statistics}
  \centering
  \begin{tabular}{lllll}
    \toprule
    \cmidrule(r){1-1}
    Property & min.  & max. & mean & std.\\
    \midrule
    Sampled Conformers & 1 & 30 & 13.8 & 7.4  \\
    Heavy Atom Count & 10 & 32 & 17.8 & 3.35 \\
    Molecular Weight & 130.0 & 507.9 & 250.5 & 53.0 \\
    Rotatable Bonds & 0  & 1 & 0.99 & 0.09   \\
    \bottomrule
  \end{tabular}
\end{table}

\section{Appendix B: Neural Network Architectures and Random Forest Baselines}

We implemented all experiments in Python using PyTorch 1.5 \citep{Paszke2019-fn} and PyTorch Geometric \citep{Fey2019-ki}.

Of the 1,157 examples above, we randomly split our data into a training set (500), validation set (200), and test set (457). We trained our networks as various subsets of the training set (n = 100, 500) for the data shown in the table above.

We trained all neural networks up to 200 epochs using the Adam optimizer (learning rate = $1 \times 10^{-3}$ to $1 \times 10^{-4}$) and a batch size ranging from 1 – 16, using the early stopping criterion based on the validation set described above. We found that a small batch size typically improved training speeds, but the increased variability occasionally led to stalled training. The model architecture and hidden dimensions are specified in Appendex Table \ref{nns}. All networks use three layers of graph featurization followed by a single attention aggregation step. The entire network is trained on binary labels ($\bf{0/1}$) for each set of conformers. 

All random forest classifiers were trained using scikit-learn, using ensembles of 100 trees. The biaryl ligands were represented using RDKit's Morgan Fingerprints (ECFP4), with the radius set to 2 and a bit vector length of 128.

\begin{table}
  \caption{Embedding-based neural network model architecture}
  \label{nns}
  \centering
  \begin{tabular}{ll}
    \toprule
    \cmidrule(r){1-1}
    Layer & Description\\
    \midrule
    Graph Featurizer &  NNConv, iterations = 3, $h_{dim} = 16$\\
    Set-Aggregator & Feed-Forward Attention $h_{dim} = 128$ $\rightarrow$ 1 \\
    Fully-Connected Layer & $h_{dim} = 16$ $\rightarrow $ 1 + Sigmoid \\
    \bottomrule
  \end{tabular}
\end{table}

\end{document}